\documentclass[acmtog]{acmart}

\AtBeginDocument{%
  }

\copyrightyear{2025}
\acmYear{2025}
\setcopyright{cc}
\setcctype{by}
\acmConference[SIGGRAPH Conference Papers '25]{Special Interest Group on Computer Graphics and Interactive Techniques Conference Conference Papers }{August 10--14, 2025}{Vancouver, BC, Canada}
\acmBooktitle{Special Interest Group on Computer Graphics and Interactive Techniques Conference Conference Papers (SIGGRAPH Conference Papers '25), August 10--14, 2025, Vancouver, BC, Canada}
\acmDOI{10.1145/3721238.3730636}
\acmISBN{979-8-4007-1540-2/2025/08}

\usepackage{multirow} 

\begin{document}

\title{3D Stylization via Large Reconstruction Model}

\author{Ipek Oztas}
\email{ipek.oztas@bilkent.edu.tr}
\affiliation{%
  \institution{Bilkent University}
  \country{Turkey}
}

\author{Duygu Ceylan}
\affiliation{%
  \institution{Adobe Research}
  \country{United Kingdom}
}

\author{Aysegul Dundar}
\affiliation{%
  \institution{Bilkent University}
  \country{Turkey}
}

\begin{abstract}
With the growing success of text or image guided 3D generators, users demand more control over the generation process, appearance stylization being one of them. Given a reference image, this requires adapting the appearance of a generated 3D asset to reflect the visual style of the reference while maintaining visual consistency from multiple viewpoints. To tackle this problem, we draw inspiration from the success of 2D stylization methods that leverage the attention mechanisms in large image generation models to capture and transfer visual style. In particular, we probe if large reconstruction models, commonly used in the context of 3D generation, has a similar capability. We discover that the certain attention blocks in these models capture the appearance specific features. By injecting features from a visual style image to such blocks, we develop a simple yet effective 3D appearance stylization method. 
Our method does not require training or test time optimization. Through both quantitative and qualitative evaluations, we demonstrate that our approach achieves superior results in terms of 3D appearance stylization, significantly improving efficiency while maintaining high-quality visual outcomes. Code and models are available via our project website: \href{https://github.com/ipekoztas/3D-Stylization-LRM}{\textbf{{https://github.com/ipekoztas/3D-Stylization-LRM}}}.

\end{abstract}




\begin{teaserfigure}
    \centering
    \includegraphics[width=0.99\textwidth]{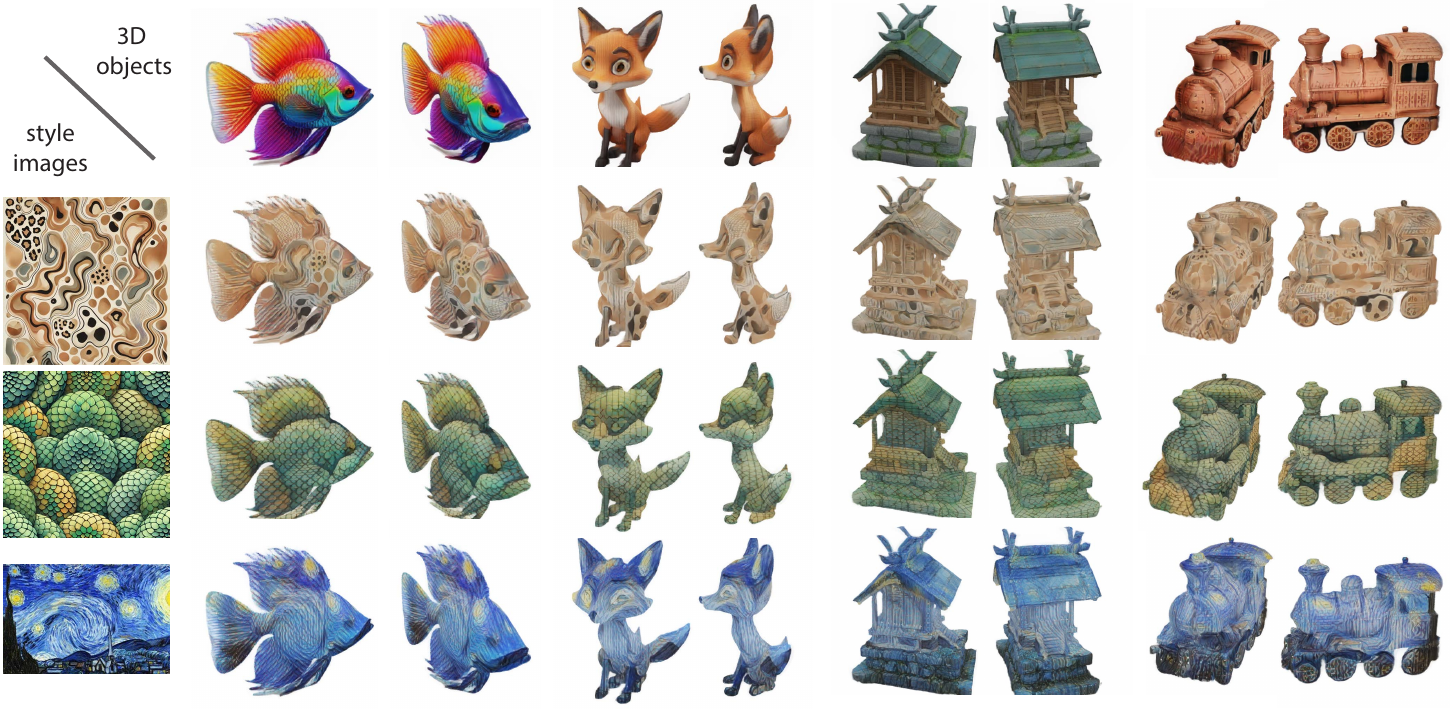}
    \caption{Given a 3D neural radiance field reconstructed by a large reconstruction model (top row), our approach enables the visual stylization based on a target style image (left column). Our method does not require any fine-tuning or test-time optimization.}
    \label{fig:teaser}
\end{teaserfigure}

\maketitle

\section{Introduction} \label{intro}

In recent years, there has been notable success in 3D generation methods that can create 3D objects from as little input as possible, such as a text prompt or a single image ~\cite{hong2023lrm, gslrm2024, li2023instant3d, liu2024one, liu2023zero, wei2024meshlrm, xu2024instantmesh}. With such powerful generators in place, users demand more control over the generation process, such as stylizing the appearance of generated 3D content. Stylizing 3D objects traditionally requires a unique blend of creativity and technical skill, especially when adapting a specific visual style to a three-dimensional form. For example, taking a reference image, such as one painted by Monet, and applying that distinctive style to a 3D generated object introduces several challenges. Unlike in 2D, where the artist can directly control the application of color, texture, and brush strokes, 3D stylization must consider how appearance and texture details should evolve with the underlying surface details. This is necessary to preserve the visual essence of the reference while generating a coherent and aesthetically pleasing appearance from every viewing angle.

In the context of 3D stylization, typically radiance fields~\cite{mildenhall2021nerf} are explored. Given a set of multi-view images of a 3D object and a reference style image, a complex optimization has proven effective in rendering realistic 3D scenes and applying the desired visual style \cite{zhang2022arf, liu2023stylerf, sun2024nerfeditor}. However, being a per-object optimization process, this is computationally expensive and time consuming. This hinders the adoption of such methods in conjunction with 3D generators where an interactive generation and stylization pipeline is desired.

To tackle the aforementioned challenge, we are inspired by the progression of 2D stylization methods where deep learning made it possible to automate the stylization of images.  
More specifically, diffusion-based large image generators, such as Stable Diffusion \cite{rombach2022high} and DALL·E \cite{ramesh2021zero}, not only offer high-quality image generation, but also complex style transfer capabilities. A key feature of these models is the use of attention mechanisms, which allow them to focus on the most relevant content and style features, enabling more nuanced and precise stylization \cite{jeong2024visual, hertz2024style, chung2024style}. Inspired by the natural ability of image generators in the stylization task, we ask if a similar capability also emerges in the context of 3D generators where similar attention mechanisms are adapted in 3D reconstruction models. 

Our goal in this work is to enable appearance stylization in the context of 3D generation by exploiting the prior knowledge of large reconstruction models \cite{hong2023lrm, li2023instant3d, xu2024instantmesh} that use a sparse set of multi-view input images to generate 3D models. 
Specifically, we investigate the architecture of these models and discover that the various attention mechanisms that capture the relation between the 3D form and input images implicitly model geometry and appearance specific features. By injecting features from a reference style image to the relevant attention modules, we show that it is possible to achieve high-quality 3D stylization without any training or per-asset optimization. We also show that the degree of stylization can be controlled by blending attention outputs with respect to original and style reference images. We evaluate our approach on a wide variety of 3D objects and style reference images and compare to several 2D and 3D stylization baselines. We show both quantitative and qualitative results that demonstrate our method achieves superior performance.

The main contributions of this paper are as follows:
\begin{itemize}
    \item We introduce the use of large pre-trained reconstruction models for 3D stylization, providing a novel approach to transferring artistic styles to 3D objects.
    \item Our method achieves 3D stylization without the need for training or test-time optimization. This makes our approach practical for interactive applications. 
   \item We provide a comprehensive evaluation with quantitative and qualitative results that demonstrate our method achieves superior 3D stylization performance compared to existing approaches.
\end{itemize}
\section{Related Work} \label{related_work}

\textbf{3D Reconstruction and Generation.} 3D reconstruction of objects from images has long been a key research challenge. Early approaches focused on training encoder-decoder architectures to capture both shape and texture, often with networks trained for specific object categories, such as birds or cars~\cite{kanazawa2018learning, chen2019learning, goel2020shape, li2020self, henderson2020leveraging, bhattad2021view, dundar2023fine, dundar2023progressive}. Recently, large reconstruction models (LRMs) \cite{hong2023lrm} have been proposed, capable of reconstructing arbitrary objects in real-world settings. This is made possible by their scalable transformer-based architecture and training on vast datasets containing around 1 million objects. Such methods have been used to output 3D representations in the form of neural radiance fields (NeRF)~\cite{hong2023lrm}, Gaussian splats~\cite{gslrm2024} \cite{tang2024lgm} \cite{xu2024grm}, and meshes~\cite{wei2024meshlrm}. 

In a different area of research, pretrained text-based image generating diffusion models \cite{rombach2022high} have been fine-tuned for novel view synthesis of objects \cite{liu2023zero, liu2024one, shi2023zero123++}. Although these methods lack a dedicated rendering mechanism to enforce strict multi-view consistency, their foundation in well-trained diffusion models allows them to produce outputs with impressive consistency and high quality.

Additionally, multi-view image generation using foundation video models has shown promise, as demonstrated by works such as V3D \cite{chen2024v3d}, Cat3D \cite{chen2024v3d}, and ViewCrafter \cite{yu2024viewcrafter}.

Finally, and most closely related to our work, are sparse-view large reconstruction models (LRMs) that are integrated with multi-view diffusion models to utilize sparse multi-view inputs, improving the quality of object reconstruction \cite{li2023instant3d, xu2024instantmesh}. Given an input image, the process begins by generating consistent multi-view images using multi-view diffusion models, which are then fed into the sparse-view large reconstruction models.
The LRM takes image embeddings from multiple views, which leads to higher-quality reconstructions.
Our work analyzes the content and appearance disentanglement of such models to build a 3D visual stylization method on top of them.

\textbf{2D Visual Style Transfer.} Transferring the style of one image to the content of another is a widely studied research topic. Early approaches relied on pretrained VGG classification networks to extract style and content features from images \cite{gatys2016image, lai2017deep, li2017universal, lu2019closed, zhang2022domain}. More recently, with the rise of diffusion models, pretrained diffusion models have been explored for this task.
Initially, methods propose fine-tuning diffusion models on a dataset with a shared style \cite{gal2022image, ruiz2023dreambooth}, enabling the generation of images in that specific style. Adapters for diffusion models are also trained \cite{ye2023ip, mou2024t2i, zhao2024uni, wang2023styleadapter} to allow the stylization of any input image \cite{yildirim2025md}, regardless of the style. 
B-Lora \cite{frenkel2025implicit} trains the LoRA weights for two specific blocks to separate the style and content components of a single image. This approach identifies particular layers within the diffusion model that encode style and content independently.
While these approaches require the time and computational cost of fine-tuning, training-free methods are introduced as a more efficient alternative. 
Attention swapping is proposed for reference based stylization to generate images with different text prompts sharing a similar visual style \cite{jeong2024visual, hertz2024style}. 
Similarly, StyleID \cite{chung2024style} manipulates the features of self-attention layers to achieve stylization of a content image.
InstantStyle \cite{wang2024instantstyle} builds upon the ideas of B-Lora, but instead of tuning LoRA weights for the content and style blocks, it provides CLIP image features to these layers to control the style. Alaluf et al.~\cite{alaluf2024cross} propose to inject features from a style image into attention layers of a diffusion model to enable cross image attention for appearance transfer. Our work is inspired by such 2D methods and aims to extend them to 3D. To the best of our knowledge, we are the first to investigate the use of attention mechanisms in large reconstruction models, to enable the transfer of visual style from a reference style image to a 3D reconstructed object.

\textbf{3D Visual Style Transfer.} In the context of 3D visual stylization, several approaches leverage NeRF (Neural Radiance Fields) as the underlying 3D representation \cite{zhang2022arf, liu2023stylerf, sun2024nerfeditor}. These methods typically rely on a consistent set of multi-view images and a training process designed to fit a 3D model to this image set. ARF \cite{zhang2022arf} formulates the stylization of radiance fields as an optimization problem, where images are rendered from multiple viewpoints, and the model minimizes a content loss between the rendered stylized images and the original multi-view images, alongside a style loss comparing the rendered images to the target style images. StyleRF \cite{liu2023stylerf} introduces a two-stage model: the first learns a 3D representation of novel views in a feature space, and the second applies stylization to these novel views. NerfEditor \cite{sun2024nerfeditor} integrates StyleGAN with the NeRF model, optimizing additional modules where StyleGAN guides the process through its interpretable latent space, while NeRF ensures multi-view consistency in the image generation. NeRF analogies \cite{fischer2024nerf} enable appearance style transfer between two NerF representations by leveraging DiNO features to construct a set of correspondences which are used to guide an appearance transfer optimization. All of these methods focus on a per-asset optimization process. In contrast, our method leverages the LRMs in a training and optimization free setup to transfer the appearance of a style image to a 3D radiance field.

With the fast adoption of Gaussian splats as a 3D representation, most recent work focuses on stylizing such representation. Kovacs et al.~\cite{gstyle} present a per-asset optimization to optimize for the color of each Gaussian and perform Gaussian splitting to enable high quality results. Liu et al.~\cite{liu2023stylegaussian} avoid per-asset optimization by embedding VGG features into each Gaussian splat which are then modulated based on a target style image and decoded back to stylized colors. The faster stylization process comes with the slightly blurry stylization results. 

\section{Method} \label{method}

\begin{figure}[t]
   \centering
\includegraphics[width=1.0\linewidth]{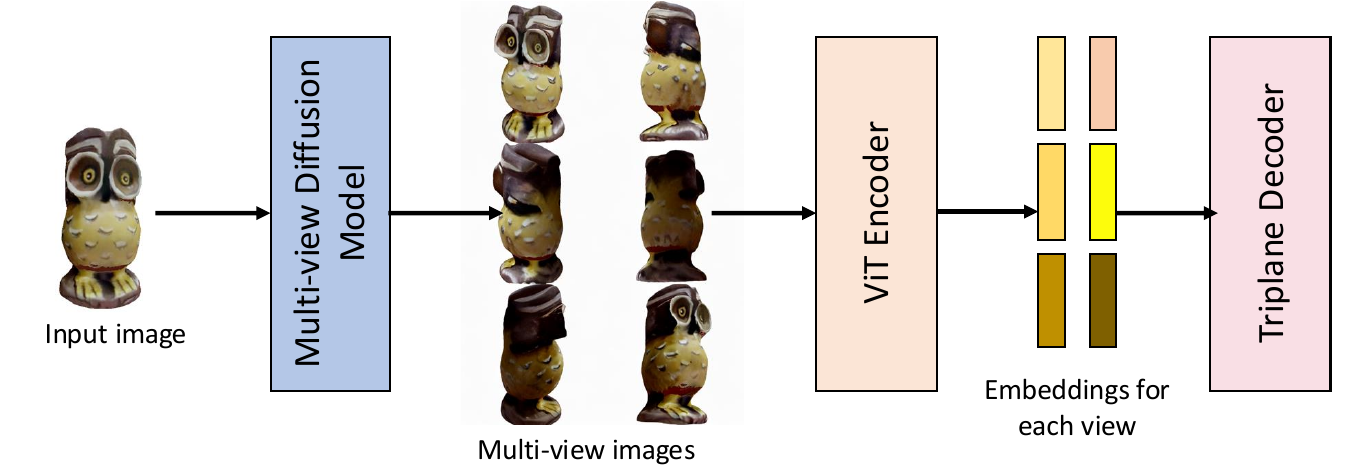}
   \caption{InstantMesh provides a two stage pipeline for single-image based reconstruction. First a multi-view diffusion model is used to generate images of the object from multiple views. Embedding of such views are then provided to a triplane decoder to generate 3D object.}
\label{fig:instant_mesh}
\end{figure}

Our work builds upon InstantMesh \cite{xu2024instantmesh} which is a framework to reconstruct 3D objects from a single image. It consists of a multiview generator and a large reconstruction model as shown in Figure~\ref{fig:instant_mesh}. Here, we first provide an overview of InstantMesh followed by a detailed presentation of our method.

\begin{figure*}[t]
    \centering
\includegraphics[width=1.0\linewidth]{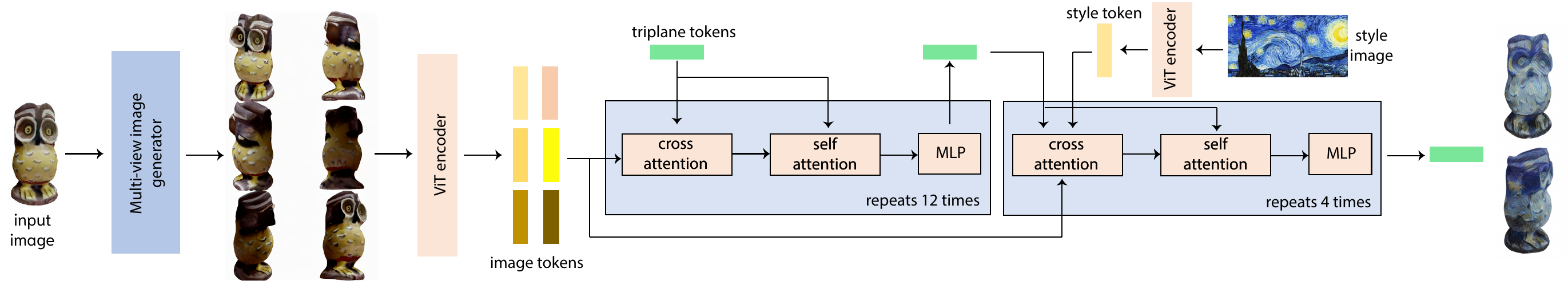}
    \caption{Overview of our method. We build upon InstantMesh which takes an input image of an object and generates 6 sparse view outputs, which are then processed by a ViT encoder. A transformer based reconstructor generates a triplane where cross attention is performed between triplane and image tokens. Given a style image, we also encode it with the ViT encoder. In the latter transformer blocks, we perform cross-attention with respect to the style image tokens and blend it with the original cross attention output. This training and optimization free approach results in high-quality 3D visual stylization.}
\label{fig:method}
\end{figure*}

\subsection{Overview of InstantMesh}
 Given an input image of an object, first a multi-view diffusion model, specifically Zero123++ \cite{shi2023zero123++}, is utilized to synthesize 6 novel views of the object at pre-defined viewpoints. These views are then given to a reconstructor to generate a 3D object represented as triplanes. The reconstructor is based on the LRM framework \cite{hong2023lrm} and is a large transformer architecture with 16 layers, each comprising a sequence of cross-attention, self-attention, and MLP blocks. It consists of a set of learnable triplane features with positional embedding, which are then transformed through the feedforward architecture into a 3D form. The features for cross-attention are derived from the embeddings of the 6 generated multi-view images extracted by a vision transformer (ViT). 
The triplane is composed of three axis-aligned planes $T_{XY}, T_{YZ}, T_{XZ}$ and can be rendered via a volumetric renderer after sampling points in 3D and further encoding them via an MLP.

%

\textbf{Attention Mechanism.}
The triplane decoder leverages a transformer architecture that is built upon the attention mechanism. Attention can be either self-attention or cross-attention, depending on the type of inputs it processes. The mechanism organizes the input features into three components: query (Q), key (K), and value (V). While the query represents what the model is currently focusing on, the key encodes the content of the input elements, and the value holds the actual information to be used in the output. During the attention process, the model computes a score for each query and key pair, typically using a dot product. These scores are then normalized using a softmax function to create attention weights. The weighted sum of the value vectors produces the final output as given in Eq. \ref{eq:attn}.

\begin{equation}
\text{Attention}(Q, K, V) = \text{softmax}\left( \frac{QK^T}{\sqrt{d_k}} \right) V
\label{eq:attn}
\end{equation}

In self-attention, all the query, key, and value features come from the same input sequence, but are projected into different spaces via learnable linear transformations. Cross-attention works similarly to self-attention, with the difference that the keys and values come from a different source than the queries. In the context of InstantMesh, the triplane features (i.e., triplane tokens) attend to each other within the self-attention modules. In cross-attention, however, the keys and values are derived from the ViT embeddings of the multi-view images (i.e., image tokens), while the query is computed from the feed-forward triplane features. 
More specifically, the queries capture the semantic meaning of each spatial 3D location, while the keys, in turn, provide context for each query. 

\subsection{3D Stylization with InstantMesh}

As discussed, the cross attention captures the relevance of different parts of input images for a given spatial 3D position and hence determines its content including both its geometry and the appearance. We hypothesize that while the earlier transformer blocks are crucial for determining the geometry (shape and structure), the later blocks focus on the appearance of the reconstructed objects. Hence, to achieve 3D stylization, we introduce a second source of query and key features, to the latter cross attention modules of the triplane decoder as shown in Fig. \ref{fig:method}. 
Given a reference style image, we first encode it using the ViT Encoder similar to the original multi-view images. 
We then incorporate the attention output with respect to the style image alongside the original multi-view images as given in Eq. \ref{eq:sty_alpha}.

\begin{multline}
\text{Attention}(Q, K, V) =    \text{softmax}\left( \frac{Q_{in}K_{style}^T}{\sqrt{d_k}} \right) V_{style} * \alpha \\
+ \text{softmax}\left( \frac{Q_{in}K_{content}^T}{\sqrt{d_k}} \right) V_{content} * (1-\alpha) 
\label{eq:sty_alpha}
\end{multline}

where $\alpha$ is a blending weight to control the degree of stylization. $K_{content}$ and $V_{content}$ denote the key and value features from the original multi-view images whereas $K_{style}$ and $V_{style}$ are the key and value features of the style image. The incorporation of the cross attention with respect to the style reference image enables the triplane decoder to correlate the 3D features with the visual stylistic elements of the reference image. Enabling style-based cross attention only in the latter transformer blocks ensures that the model first builds a solid geometric foundation before progressively incorporating stylistic details. The resulting model preserves the underlying geometry while integrating the desired visual style, resulting in a well-balanced and high-quality stylization.

\begin{figure*}[t]
    \centering
\includegraphics[width=0.8\linewidth]{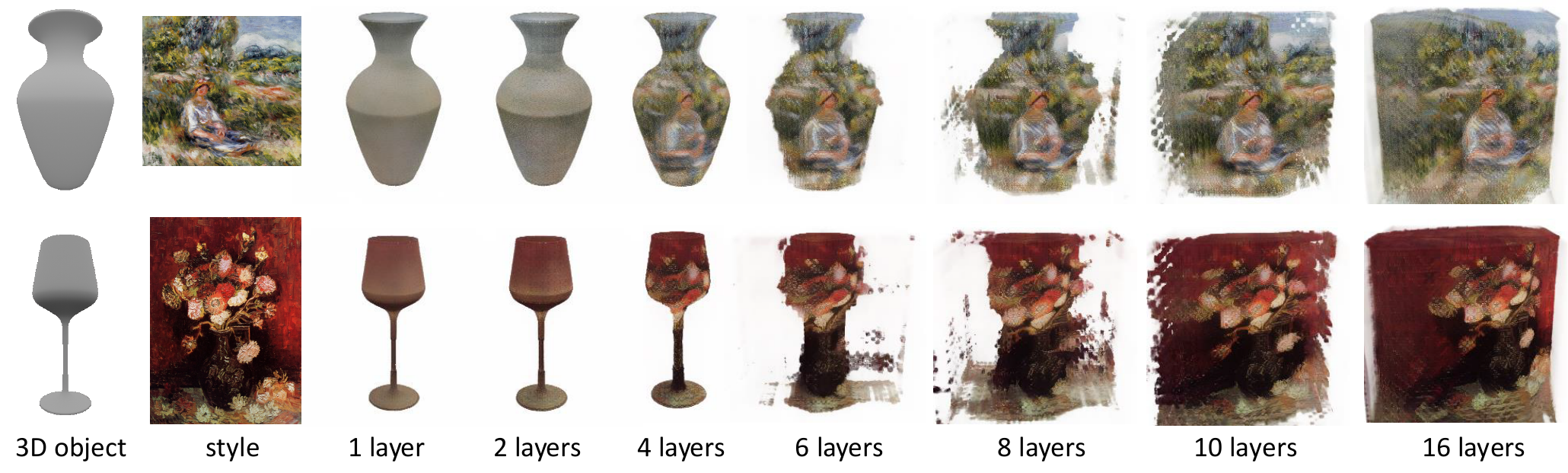}
    \caption{Qualitative results for using style image ViT-encoded embeddings in the cross-attention layers across the last 1, 2, 4, 6, 8, 10, and all 16 layers of the triplane decoder. We use the last 4 layers in our experiments which provide a good balance between visual stylization and content fidelity.}
\label{fig:ablation_layers}
\end{figure*}

\begin{table*}[t!]
    \centering
    \begin{tabular}{cccccccc}
        \toprule
        Layers: & 1 & 2 & 4 & 6 & 8 & 10 & 16 \\
        \midrule
        CD:     & 0.00018 & 0.00077 & 0.00594 & 0.04281 & 0.16933 & 0.23368 & 0.26292 \\
        \bottomrule
    \end{tabular}
    \caption{CD values across different layers.}
    \label{tab:cd_values_horizontal}
\end{table*}

In Fig. \ref{fig:ablation_layers}, we illustrate the impact of providing the embeddings of the style image to varying numbers of transformer blocks in the triplane decoder. As shown, when style reference features are injected into more blocks, its content starts interfering with the underlying 3D geometry and results in suboptimal results. This observation is consistent with previous diffusion based 2D style transfer methods~\cite{alaluf2024cross}, which inject style image features only to later blocks of the diffusion model. 
Table \ref{tab:cd_values_horizontal} presents the computed Chamfer Distance (CD) scores from our ablation study on the number of layers used for style token injection. Based on this observation, we enable cross attention with style reference image features only in the last 4 transformer blocks.

%

\begin{figure}[t]
    \centering
\includegraphics[width=0.8\linewidth]{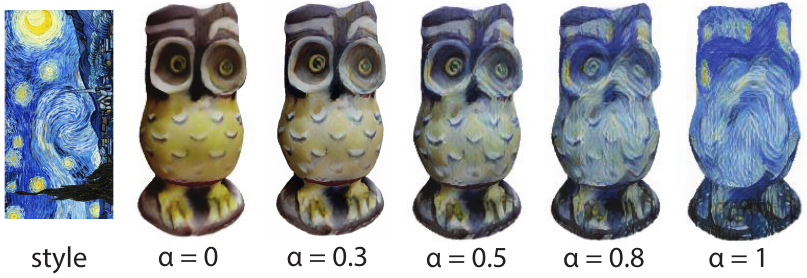}
    \caption{Our method blends the cross attention output with respect to original and style images based on a blending factor $\alpha$. We set $\alpha=0.8$ in our experiments.}
\label{fig:abl_alpha}
\end{figure}

While we observe a separation of how geometry and appearance features are handled across different transformer blocks, we also note that for certain type of 3D objects, certain intricate details might be better preserved in the appearance rather than the geometry. In such cases, instead of overriding cross attention output with attention results with respect to the style reference image only, we choose to blend with attention with respect to the original multi-view images using a blending weight $\alpha$. As illustrated in Fig. \ref{fig:abl_alpha}, the intricate details of the owl's eyes are better preserved in the appearance rather than the geometry, and such details are preferred to be maintained during stylization. In our experiments, we empirically set $\alpha=0.8$ to provide a good balance. 
%
%
%
\section{Experiments} \label{experiments
}


\begin{figure*}[t]
    \centering
\includegraphics[width=1.0\linewidth]{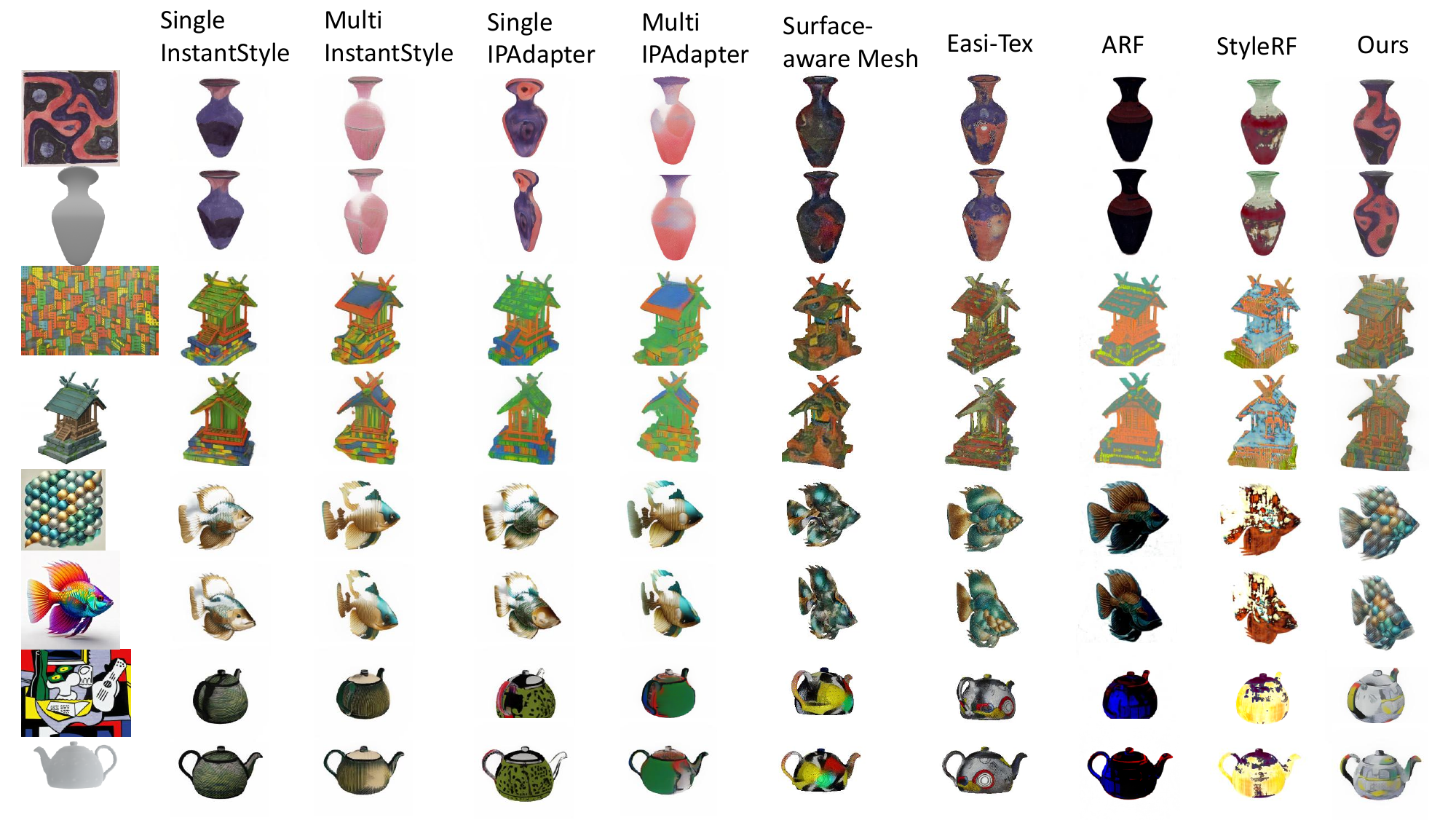}
    \caption{Visual results of our method and competing methods.}
\label{fig:comparisons}
\end{figure*}

\noindent \textbf{Dataset.} 
We evaluate our method on a wide variety of 3D reconstructed objects. Specifically, we use renderings of objects from Google Scanned Objects (GSO) \cite{downs2022google} and OmniObject3D (Omni3D) \cite{wu2023omniobject3d}. We also generate images using prompts provided by previous work~\cite{li2023instant3d} and a text-to-image generation model\footnote{https://firefly.adobe.com/}. 
We collect a set of style reference images from Wikiart \cite{tan2018improved} as well as generating them from text prompts. For quantitative evaluation, we use $15$ 3D objects where we apply $2$ different style images selected randomly from a set of $30$ style images.

\noindent \textbf{Metrics.} To evaluate style fidelity, following \cite{alaluf2024cross}, we assess the similarity of Gram matrices. This is a concept adopted from the neural style transfer literature \cite{gatys2016image}, which suggests that images with similar styles tend to have similar Gram matrices. Specifically, we compute the L2 distance between the Gram matrices of the appearance image and the generated image, using an ImageNet pretrained VGG19 network to extract feature representations.

To evaluate content fidelity and ensure that the visual stylization does not degrade 3D geometry quality, we also report the Chamfer distance between the original 3D object and the stylization result using a similar pipeline as in InstantMesh~\cite{xu2024instantmesh}. 
Specifically, each 3D object is normalized into a unit cube and Chamfer distance is computed based on 16K points sampled from the surface uniformly.

\noindent \textbf{Baselines.}
We establish several baselines to demonstrate the effectiveness of our method. The first set of baselines involve stylizing single input images, which are then processed by InstantMesh (i.e., we first generate multi-view images given the stylized single view and perform 3D reconstruction). Given the recent surge in single-image stylization methods, we conduct experiments using InstantStyle \cite{wang2024instantstyle} and IpAdapter \cite{ye2023ip}, both of which have shown to produce high-quality image stylization results. We also compare our approach against additional techniques—namely, StyleAlign \cite{hertz2024style}, Cross-Image Attention \cite{alaluf2024cross}, AdaIN \cite{huang2017adain}, and Attention Interpolation \cite{he2024aid}—to further validate its performance.

In the next set of comparisons, similar to Style-NeRF2NeRF~\cite{Fujiwara2024}, we perform stylization on multi-view images which are provided as input to the 3D reconstructor. To ensure consistency across views, we employ the grid trick, a technique commonly used in video stylization \cite{kara2024rave} and 3D texturing \cite{ceylan2024matatlas}. Specifically, we stylize multiple views in a single image generation pass by stacking them in a $3\times2$ grid pattern. This setup allows each view in the grid to potentially attend to the features of the other views, effectively implementing the full-shared attention proposed by Style-NeRF2NeRF.
We use InstantStyle \cite{wang2024instantstyle}, IpAdapter \cite{ye2023ip} StyleAlign \cite{hertz2024style}, Cross-Image Attention \cite{alaluf2024cross}, AdaIN \cite{huang2017adain}, and Attention Interpolation \cite{he2024aid} as image stylization methods in this setup.

Additionally, compare our approach with texture synthesis and texturing methods: Surface‐Aware Mesh Texture Synthesis with Pre‐trained 2D CNNs \cite{kovacs2024meshtexturesynthesis} for image-based texture synthesis, and EASI-Tex \cite{perla2024easitex} for texturing via its IP-Adapter generalization.

Finally, we compare our approach with optimization-based 3D stylization methods, specifically StyleRF \cite{liu2023stylerf} and ARF \cite{zhang2022arf}. These methods rely on multi-view images, so we prepare a dataset for each object, rendering 100 views for training, 100 views for validation, and 120 views for test using ground truth camera parameters.


\begin{table}[t!]
\caption{Comparisons with competing methods. "single" refers to stylization using a single input image, while "multi" refers to stylization based on multi-view images.}
\centering
\footnotesize
\begin{tabular}{|l|c|c|c|c|c|}
\toprule
\multirow{2}{*}{Method} & {Style}  & \multirow{2}{*}{CD$\downarrow$} &  \multirow{2}{*}{F Score$\uparrow$} &\multirow{2}{*}{Run-time} \\
&  Fidelity $\downarrow$ & &&  \\
\toprule
Single - InstantStyle & 41.915 & 0.01380 & 0.9634& 5.9 min \\
Single - IPAdapter & 38.550 & 0.01567 &0.9476 & 3.2 min\\
Single - StyleAlign & 45.346 & 0.02402& 0.91915& 2.8 min\\
Single - Cross-Image Attn & 39.721 & 0.03042&0.90508& 4.3 min\\
Single - AdaIN & 37.980&0.04331 &0.83786&2.8 min \\
Single - Attention Interpolation & 40.304 & 0.07630&0.73042& 6 min\\

\hline
Multi - InstantStyle & 40.393 &  0.01366& 0.9669 & 4.7 min\\
Multi - IPAdapter & 37.627 & 0.01366 &0.9580 &3.2 min\\
Multi - StyleAlign & 44.456 & 0.00744 &0.95635&2.8 min \\
Multi - Cross-Image Attn & 44.834 & 0.02006& 0.9333 &4.3 min \\
Multi - AdaIN & 37.172 & 0.00692& 0.96030&2.8 min \\
Multi - Attention Interpolation & 45.353 &0.01688 &0.94476&6 min \\

\hline
ARF & 37.901 & 0.00060 & \textbf{0.99999} & 1.5 h \\
StyleRF & 42.960 & 0.00072 &0.99974& 2 h \\
\hline
EASI-TEX & 44.630 &\textbf{ 0.00032} &\textbf{0.99999}&  37 min\\
Surface‐aware Mesh  &  43.393 &\textbf{ 0.00032}& \textbf{1.0} &  23 min\\
\hline
Ours & \textbf{34.584} & 0.00594 & 0.9755 & \textbf{35.5 sec}\\ 
\bottomrule
\end{tabular}
\label{table:comparisons}
\end{table}

\noindent \textbf{Quantitative Results.} We present quantitative results in Table \ref{table:comparisons}, which includes style fidelity score, Chamfer Distance (CD) F-Score (FS) with a threshold of 0.2, which are computed by sampling 16K points from the surface uniformly and run-time of each method. We measure the run-times on a single NVIDIA A100 GPU.

Our method achieves the best style fidelity score, significantly outperforming other approaches in maintaining the target style. For image-based stylization baselines, we do not observe a significant difference between stylizing a single view vs multi-views both of which are inferior to our approach.

In terms of content fidelity, 3D approaches including ours perform on-par and preserve the 3D form of the object.
Although Nerf-based methods yield slightly better content fidelity results than ours, they necessitate a multi-view set of images. It is important to note that EASI-Tex and Surface-aware Mesh report flawless geometry because they are provided with ground-truth geometry and are only tasked with generating the texture. Image-based methods, on the other hand, suffer from geometry artifacts. We note that for each 3D approach, we consider the non-stylized reconstructions obtained by that particular method as the original 3D object. For image based stylization methods, we treat the reconstruction by InstantMesh from the original input image as the original 3D object. 

\begin{table}[t!]
\caption{User Study. We conduct a user study where we ask the participants to choose which of the presented results (obtained via baselines and our method) they prefer with respect to (i) overall quality, (ii) style fidelity, and (iii) content fidelity. In all categories, our method is preferred by a significant margin.}
\centering
\footnotesize
\begin{tabular}{|l|c|c|c|c|c|}
\toprule
\multirow{2}{*}{Method} &Overall  & Style  & Content   \\
&Quality  &Fidelity&Fidelity  \\
\toprule
Single InstantStyle & 10.73 &5.82 & 15.91\\
Single IPAdapter  & 4.07 & 4.81 & 5.18\\
\hline
Multi InstantStyle  & 7.03 & 3.7 & 12.21 \\
Multi IPAdapter & 3.33 & 7.03 & 3.33 \\
\hline
ARF &2.96 & 1.85 & 11.84 \\
StyleRF  & 1.11 & 0 &4.44\\
\hline
Ours & \textbf{62.99} & \textbf{67.81} & \textbf{59.27}\\ 
\bottomrule
\end{tabular}
\label{table:user_study}
\end{table}

Regarding computational time, our method is the fastest since once original multi-view images of an object is generated, it involves running only the 3D reconstruction for each style image. In contrast, image based stylization methods (both single and multi-view) also need to run an image generation pass adding an additional cost. In case of single image stylization, the first generation stylizes a single view and the second generation produces multi-views. In case of multi-view stylization, a pass of stylization is performed on the grid image of multiple views. NeRF-based models are significantly slower due to the need for the per-asset optimization. We note that we only report the time needed for the optimization that uses the style image excluding the time needed to reconstruct the original 3D object (which takes an additional ~30 minutes). StyleRF also requires an additional feature training setup given the original 3D radiance field which takes around ~1.5 hours. Texture generation models include maintaining multi-view consistency and the slow runtime caused by the iterative and sequential nature of texture generation across different views. Both EASI-Tex and Surface-aware Mesh exhibit significantly slower runtimes, taking over 25 minutes compared to just 35 seconds for ours.

\begin{figure*}[t]
    \centering
\includegraphics[width=0.95\linewidth]{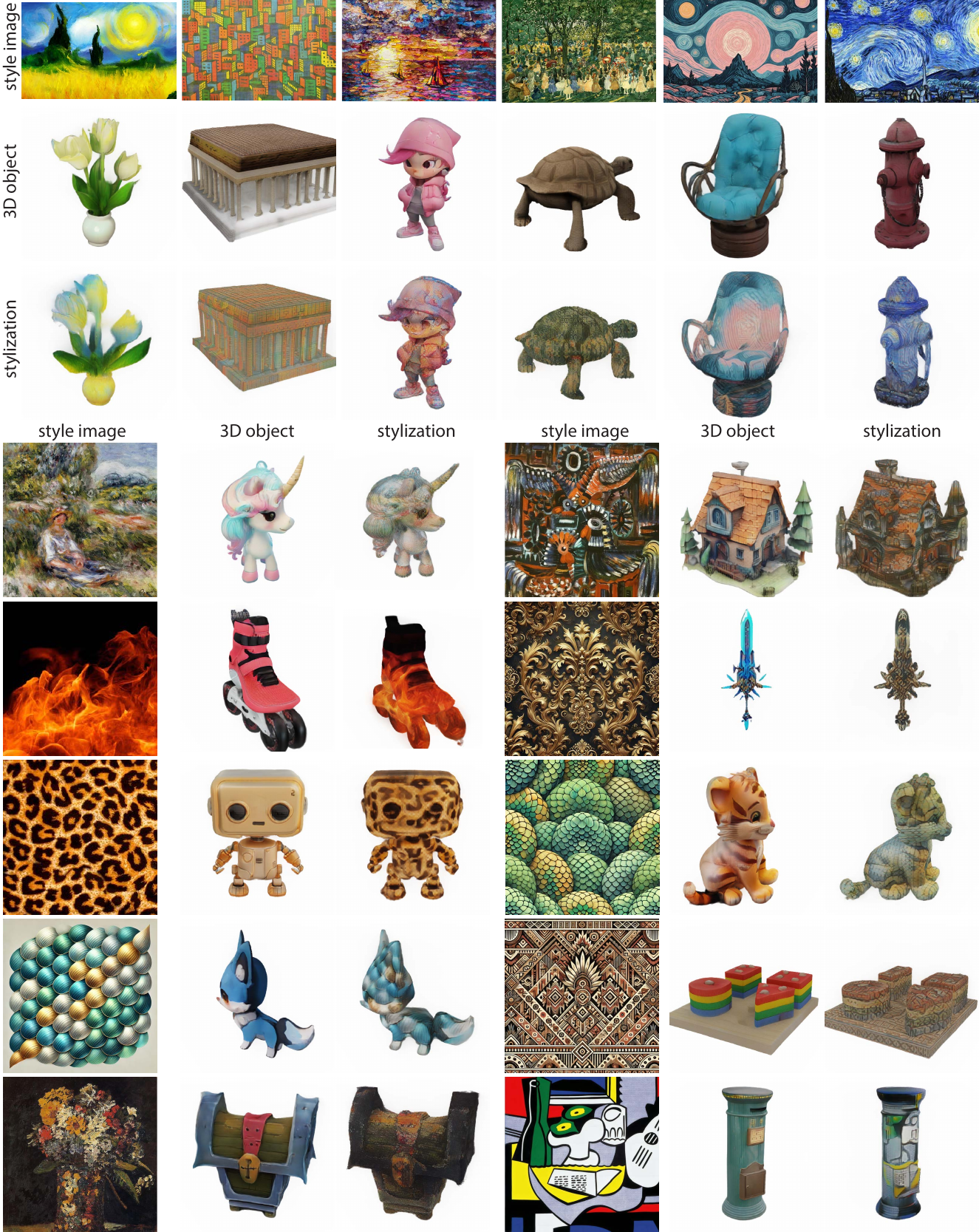}
    \caption{We provide more visual results as style image, 3D object, and stylization triplets.}
\label{fig:more_results}
\end{figure*}

\begin{figure*}[t]
    \centering
\includegraphics[width=1.0\linewidth]{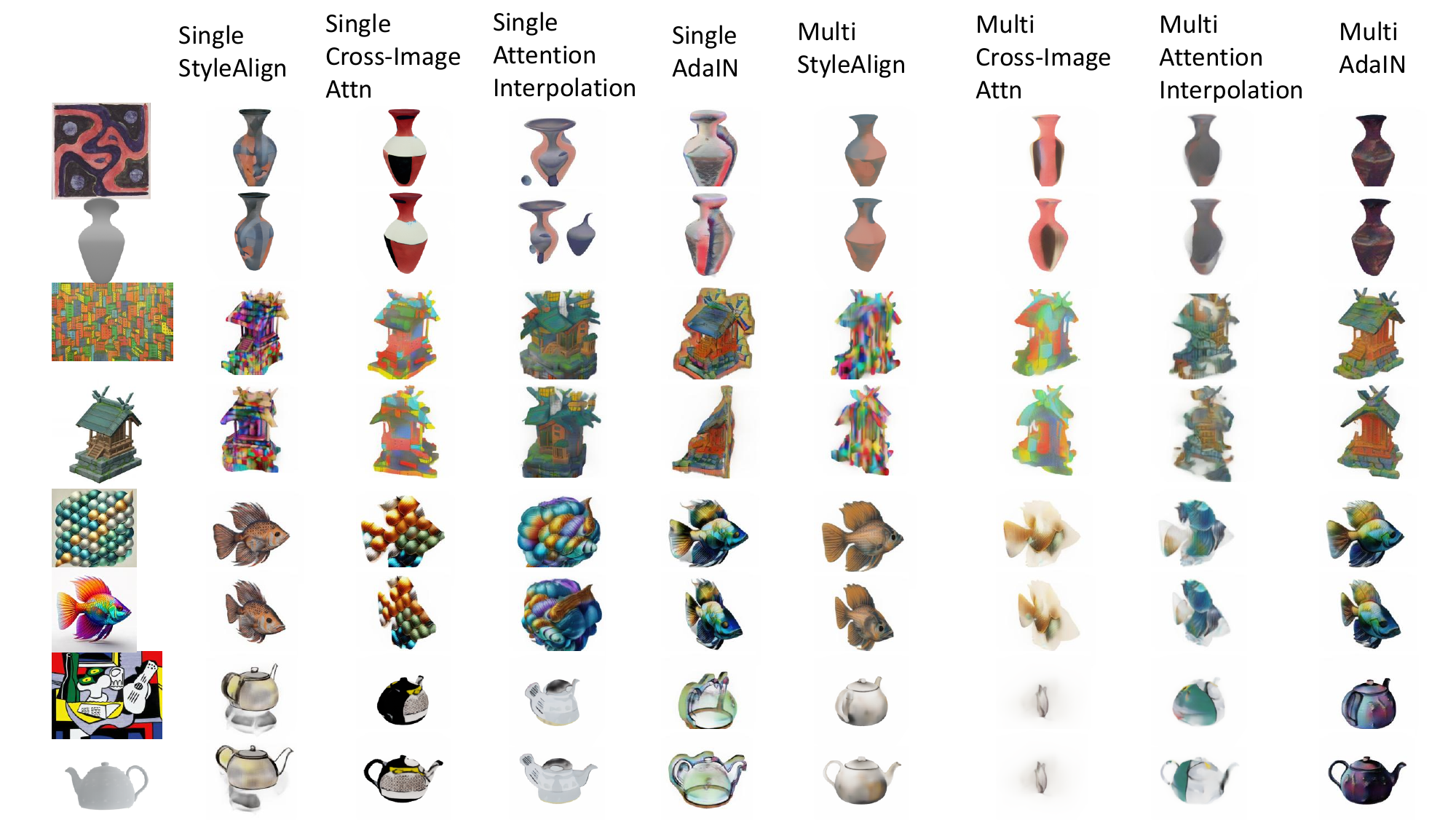}
    \caption{We provide visual results of additional competing methods.}
\label{fig:comparisons2}
\end{figure*}

\noindent \textbf{Qualitative Results.}
Fig. \ref{fig:comparisons} and Fig. \ref{fig:comparisons2} show the visual results of our method and baselines for various 3D objects and visual styles. We refer to Fig. \ref{fig:more_results} and the supplementary material for more results. We observe significant artifacts in the final 3D geometry of the results obtained by image stylization baselines. We attribute this to several factors. First, when we stylize the original single input image, the multi-view generator cannot handle such stylized images robustly and generates inconsistent multi-view images. Similarly, when stylizing multi-view images, even when stacked as a grid, stylization in the image domain does not preserve 3D consistency. Finally, when the large reconstruction model is provided with only stylized images, it may fail to accurately infer the structure of the object, resulting in flat or distorted 3D outputs (e.g., the vase).
NeRF models, while good at keeping the 3D structure intact, struggle to correctly stylize the objects, often failing to transfer the style correctly. In particular, we observe that, given darker style images, ARF tends to produce outputs with excessively dark illumination. On the other hand, StyleRF does not faithfully transfer the style of the input image; instead, it focuses on matching the color schemes of both the style and target images resulting in over-exposed color values.

\noindent \textbf{User Study.} Finally, we conduct a user study in which participants are presented with the renderings of a reconstructed 3D object, a reference style image, and the renderings of the resulting stylized object. We ask the participants to evaluate (1) how well the 3D geometry is preserved, maintaining content fidelity, (2) how well the output reflects the visual style of the reference image, and (3) the overall quality of the output. We provide our results along with the outputs from baseline methods in randomized order, and ask the participants to select the most favorable result for each of the three questions. We collect responses from $27$ participants and provide the results in Table~\ref{table:user_study}. The data reveals a strong preference for our approach over competing methods, with participants rating it highest for stylization, content fidelity, and overall quality.

\begin{figure}[t]
    \centering
\includegraphics[width=0.95\linewidth]{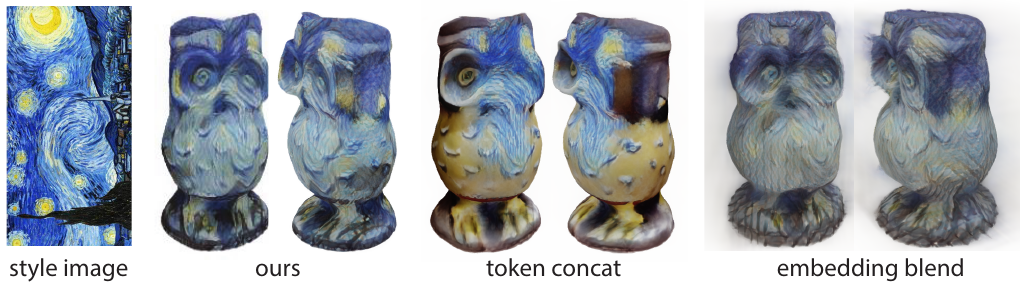}
    \caption{We compare our results to alternative design choices. First, we concatenate multi-view image and style image tokens and provide to cross attention layers. In the second alternative, we blend the embeddings of multi-view images with that of the style image before we provide to the cross attention layers. Both alternatives provide sub-optimal results.}
\label{fig:ablation-design}
\end{figure}

\begin{figure*}[t]
    \centering
\includegraphics[width=0.75\linewidth]{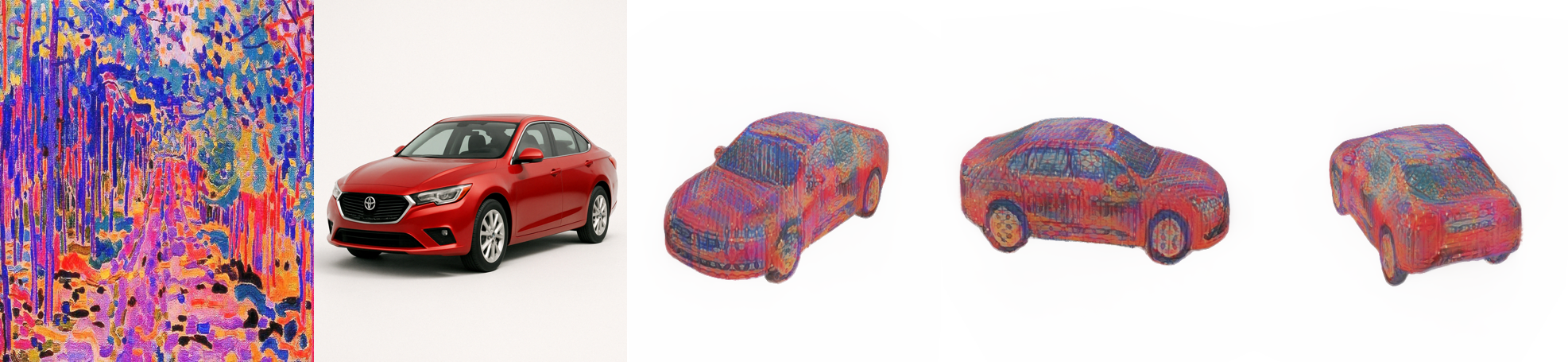}
\Description{}
    \caption{Example for limitation}
\label{fig:limitation}
\end{figure*}

\noindent \textbf{Ablations.} We ablate our design choice of computing cross attention with style image tokens with various alternatives in Fig.~\ref{fig:ablation-design}. First, we concatenate the image tokens from the original images and the style image and compute the cross attention between this larger number of tokens and the triplane tokens. As shown in the figure, in this variation, style is not transferred to the whole object consistently. Next, instead of blending the attention outputs with respect to the original and style images, we blend the ViT image embeddings of the original and style images by averaging. We then compute cross attention with respect to such blended image embeddings. This option results in degradation in 3D structure resulting in fuzzier reconstructions. Our choice of computing cross attention with respect to original and style image separately and blending the attention output as desired provides the most effective solution.
\textbf{Limitations.} Regarding limitations, while our method provides an efficient approach to stylization, it is important to acknowledge that this process may result in the abstraction or simplification of some of the finer, intricate details within the content's texture, an example is given in Fig. \ref{fig:limitation}.
Another limitation is that our work is currently influenced by the reconstruction performance of InstantMesh. However, our work will benefit from enhanced performance of large reconstruction models as extensive research continues in this direction.

\section{Conclusion} \label{conclusion}

In this paper, we introduce a novel approach to visual 3D stylization that leverages large pre-trained reconstruction models to generate 3D objects from a single image. By eliminating the need for time-consuming optimization, our method provides a simpler and more efficient alternative to traditional NeRF-based optimization techniques. Through both quantitative and qualitative evaluations, we demonstrate that our approach maintains high-quality visual outcomes. Our findings highlight the potential of this technique for real-time applications, offering a promising solution to the challenges of transferring artistic styles to 3D objects. 
While our method is capable of transferring visual styles from a reference image, we mostly keep the geometry of the 3D object intact. It is an interesting future direction to explore how to change the high frequency geometric details as well for certain styles to improve plausibility. We currently apply our model on InstantMesh for which pre-trained models are available. Extending the proposed concept to more recent approaches that reconstruct 3D Gaussian splats is another promising direction.

\textbf{Acknowledgements.} This work was supported by the BAGEP Award of the Science Academy.  We acknowledge EuroHPC Joint Undertaking for awarding the project ID EHPC-AI-2024A02-031 access to Leonardo at CINECA, Italy.


\bibliographystyle{ACM-Reference-Format}
\bibliography{references}

\appendix

\end{document}